\documentclass[letterpaper, 10 pt, conference]{ieeeconf}

\IEEEoverridecommandlockouts

\usepackage{booktabs} 
\usepackage{xcolor}
\usepackage[utf8]{inputenc}
\usepackage{float}
\usepackage{amssymb}
\usepackage{amsmath}
\usepackage{ifthen}
\usepackage{caption}
\usepackage[bookmarks=false]{hyperref}
\usepackage{url,moreverb,xspace}
\usepackage{tcolorbox}
\usepackage{array,graphicx}
\usepackage{soul}
\usepackage{balance}
\usepackage{pifont}
\usepackage{nicefrac}
\usepackage{mathtools}
\usepackage{tabularx}
\usepackage{xcolor,pifont}
\usepackage{tikz}
\usepackage{svg}
\usepackage{pdfpages}
\usepackage{array}
\usepackage{subcaption}
\usepackage{multicol}
\usepackage{multirow}
\usepackage{pgfplots}
\usepackage{makecell}
\usepackage{xcolor}
\usepackage{listings}

\lstdefinestyle{customLatex}{
    basicstyle=\ttfamily\footnotesize,
    backgroundcolor=\color{gray!10},
    keywordstyle=\color{blue},
    commentstyle=\color{gray},
    stringstyle=\color{red},
    frame=single,
    breaklines=true,
    columns=fullflexible
}
\newcommand*\colourcheck[1]{%
  \expandafter\newcommand\csname #1check\endcsname{\textcolor{#1}{\ding{52}}}%
}
\newcommand*\colourcross[1]{%
  \expandafter\newcommand\csname #1cross\endcsname{\textcolor{#1}{\ding{56}}}%
}
\usepackage[vlined, boxruled, linesnumbered] {algorithm2e}
\SetKw{KwBy}{by}
\SetKw{KwBreak}{break}
\SetKw{KwReturn}{return}
\def\HiLi{\leavevmode\rlap{\hbox to \hsize{\color{gray!35}\leaders\hrule height .8\baselineskip depth .5ex\hfill}}}

\newcommand{\company}{BMW\xspace} %
\newcommand{\carqa}{CarExpert\xspace}
\newcommand{\vehiclebrand}{BMW SUV\xspace}

\SetKwComment{Comment}{$\triangleright$\ }{}
\SetCommentSty{itshape}
\newboolean{showcomments}
\setboolean{showcomments}{true}
\ifthenelse{\boolean{showcomments}}
{\newcommand{\nb}[2] {
  \fcolorbox{black}{gray!20}{\bfseries\sffamily\scriptsize#1:}
  {\sf\small$\blacktriangleright$\textit{#2}$\blacktriangleleft$}
}
}
{\newcommand{\nb}[2]{}
}

\newcommand{\head}[1]{\noindent\textbf{#1.}}

\begin{document}

\pagenumbering{arabic} 
\pagestyle{plain}

\title{\LARGE \bf
Automated Factual Benchmarking for In-Car Conversational Systems using Large Language Models
}

\author{Rafael Giebisch$^{1}$ and Ken E. Friedl$^{2}$ and Lev Sorokin$^{3}$ and Andrea Stocco$^{4}$
\thanks{$^{1}$Rafael Giebisch is with the Technical University of Munich, Munich, Germany
        {\tt\small rafael.giebisch@tum.de}}%
\thanks{$^{2}$Ken E. Friedl is with the BMW Group, Munich, Germany
        {\tt\small ken.friedl@bmw.de}}%
\thanks{$^{3}$Lev Sorokin is with the BMW Group and the Technical University of Munich, Munich, Germany
        {\tt\small lev.sorokin@tum.de}}%
\thanks{$^{4}$Andrea Stocco is with the Technical University of Munich and fortiss GmbH, Munich, Germany
        {\tt\small andrea.stocco@tum.de}}%
}


\maketitle
\thispagestyle{empty}
\pagestyle{plain}

\begin{abstract}
In-car conversational systems bring the promise to improve the in-vehicle user experience. Modern conversational systems are based on Large Language Models (LLMs), which makes them prone to errors such as hallucinations, i.e., inaccurate, fictitious, and therefore factually incorrect information.
In this paper, we present an LLM-based methodology for the automatic factual benchmarking of in-car conversational systems. We instantiate our methodology with five LLM-based methods, leveraging ensembling techniques and diverse personae to enhance agreement and minimize hallucinations.
We use our methodology to evaluate \carqa, an in-car retrieval-augmented conversational question answering system, with respect to the factual correctness to a vehicle's manual. We produced a novel dataset specifically created for the in-car domain, and tested our methodology against an expert evaluation. Our results show that the combination of GPT-4 with the Input Output Prompting achieves over 90\% factual correctness agreement rate with expert evaluations, other than being the most efficient approach yielding an average response time of 4.5s. Our findings suggest that LLM-based testing constitutes a viable approach for the validation of conversational systems regarding their factual correctness.

\end{abstract}

\section{Introduction}\label{sec:introduction}
Researchers and companies are exploring the potential of Large Language Models (LLMs) to tackle complex tasks across diverse domains, such as content quality classification of Q\&A websites~\cite{CHAN2024100114}, code translation~\cite{10.1145/3597503.3639226}, security vulnerability detection~\cite{10.1145/3687251.3687253} and autonomous driving~\cite{10611018, cuiDrivellm24, TongLLM2024}. 

In the automotive domain, LLMs can be utilized to develop modern in-car conversational systems. These systems enhance driver and passenger experiences by enabling natural, context-aware interactions for navigation, entertainment, and vehicle control~\cite{10.1145/3242587.3242593}. These systems are expected to provide real-time information, adapt to user preferences, thereby reducing the need for manual inputs, and enhancing overall driving experience and safety~\cite{du2024proactiveinteractionsinvehicleconversational}.
However, despite their potential advantages, LLM-based in-car conversational systems also present critical development challenges in terms of factual correctness, latency, and privacy. 

\begin{figure}[t]
  \centering 
  \includegraphics[width=0.9\columnwidth]{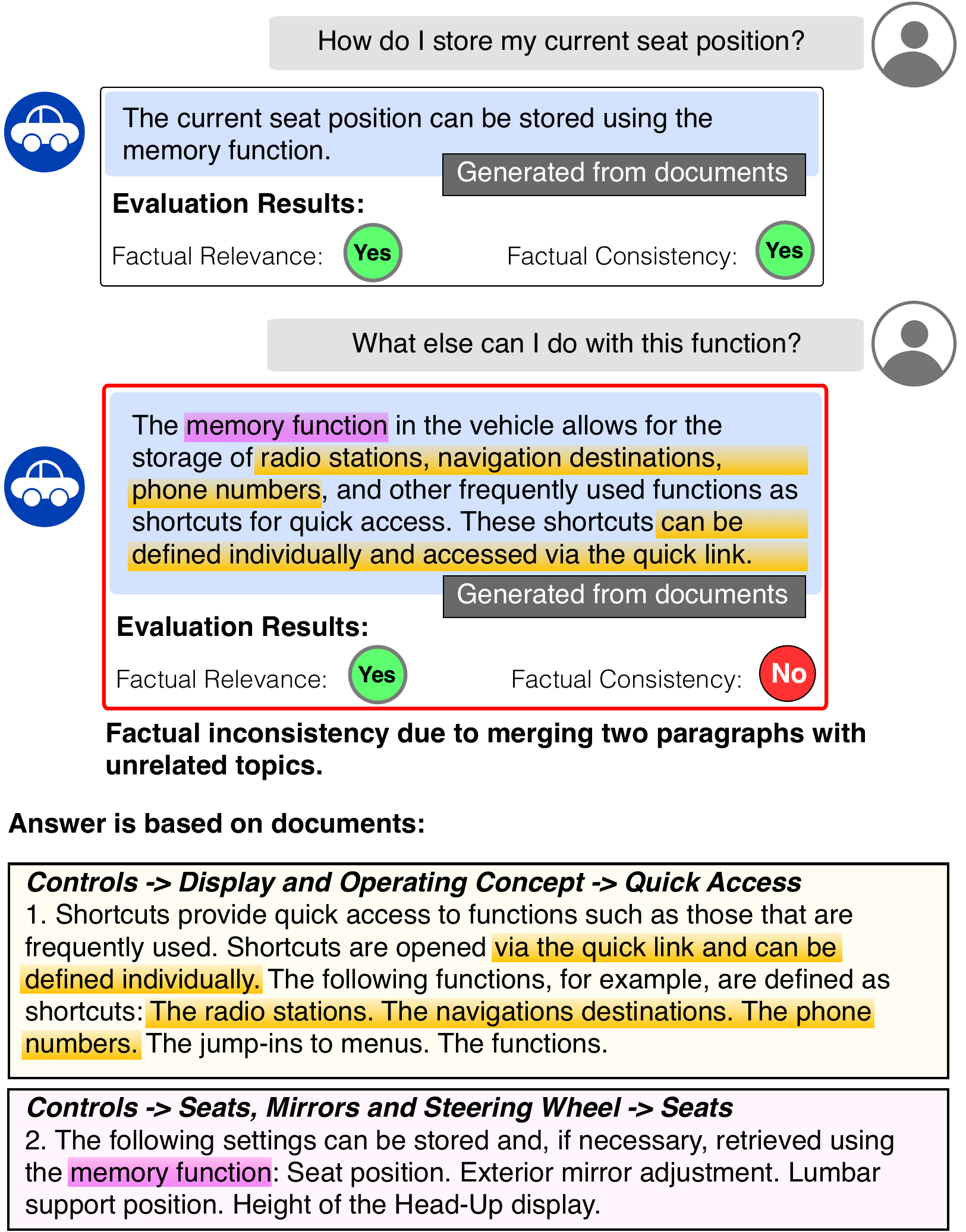}
  \caption{
  Example dialogues showing both positive (top) and negative (bottom) interactions between a user and \carqa. In the second response, \carqa fails to combine documents from the manual to provide a correct response.}
  \label{fig:dialog}
\end{figure}

In this paper, we target the automated testing of factual correctness of answers provided by a LLM-based conversational system, as it represents the vital requirement to be satisfied and thoroughly tested~\cite{2020-Riccio-EMSE} (i.e., the system should respond with factually accurate information), without
which such systems would be hardly accepted in production~\cite{Sarker2024}. 
Particularly, in this paper we focus on the evaluation of \carqa~\cite{rony-etal-2023-carexpert}, an in-car conversational system developed at \company for quality assessment. 
Among the many features, \carqa is designed to engage drivers in natural, multi-turn conversations about the vehicle and its features (\autoref{fig:dialog}). The system's responses are informed by data derived from the owner's manual, which has been parsed, annotated by domain experts, embedded, and stored in a vector database to serve as ground truth for \carqa. 

In literature, there exists several \textit{manual} approaches to help minimize hallucinations and factual inaccuracies in conversational system responses~\cite{yao2024llmlieshallucinationsbugs,banerjee2024llmshallucinateneedlive,zhang2024llmhallucinationspracticalcode}. 
However, manually debugging these defects is generally impractical for engineers due to the extensive domain knowledge and time required for a comprehensive inspection.

This paper investigates the problem of building a black-box testing framework for the automated benchmarking of the factual correctness of in-car conversational-systems, such as \carqa~\cite{rony-etal-2023-carexpert}, to reduce the manual workload for engineers. 
The black-box approach is necessary as these systems are often developed, fully or partially, by third parties.
Our approach leverages LLMs, due to their impressive performance in natural language processing, while minimizing hallucinations through techniques such as ensembling, diverse personae, and majority voting.

In our study, we evaluate the property of factual correctness using two dimensions, namely factual consistency and factual relevance. In the context of \carqa, a system response is considered as \textit{factually consistent} if it fully aligns with the information in the owner's manual. The owners' manual serves as a carefully curated dataset and source of ground truth. No additional information should be introduced, assumed, misinterpreted, taken out of context, or fabricated in the reponse to the user.
A system response is \textit{factually relevant}, if it is adequately addressing the user's question. A response can be factually consistent, but not relevant, and vice versa (see \autoref{fig:dialog}).

In this paper, we first curated a dataset of question and answer pairs for the automotive domain, where questions were generated by humans, while answers were provided by \carqa. Domain experts performed then a factual correctness evaluation on the answers.  We then selected five different types of LLMs and reasoning methods, each with custom prompt templates. Each prompt received the generated answer by \carqa and the relevant paragraphs retrieved from the manual by \carqa as input. The prompt is then passed to our framework LLMs to evaluate the degree of factual correctness.
 
We assessed the effectiveness of our framework by comparing our systems factually correctness evaluation responses with those collected using human experts. 
Additionally, we analyzed its efficiency to determine its suitability for real-time use.  

Our results show that combining multiple LLMs and methods yields over 90\% accuracy in both factual relevance and consistency. GPT-4, utilizing the Input-Output Prompting method, achieved the best tradeoff between effectiveness and efficiency, with accuracy scores of 90\% (relevance) and 92\% (concistency), and an average execution time of 4.5 seconds per request.




\begin{figure*}[t]
  \centering 
  \includegraphics[width=0.9\textwidth]{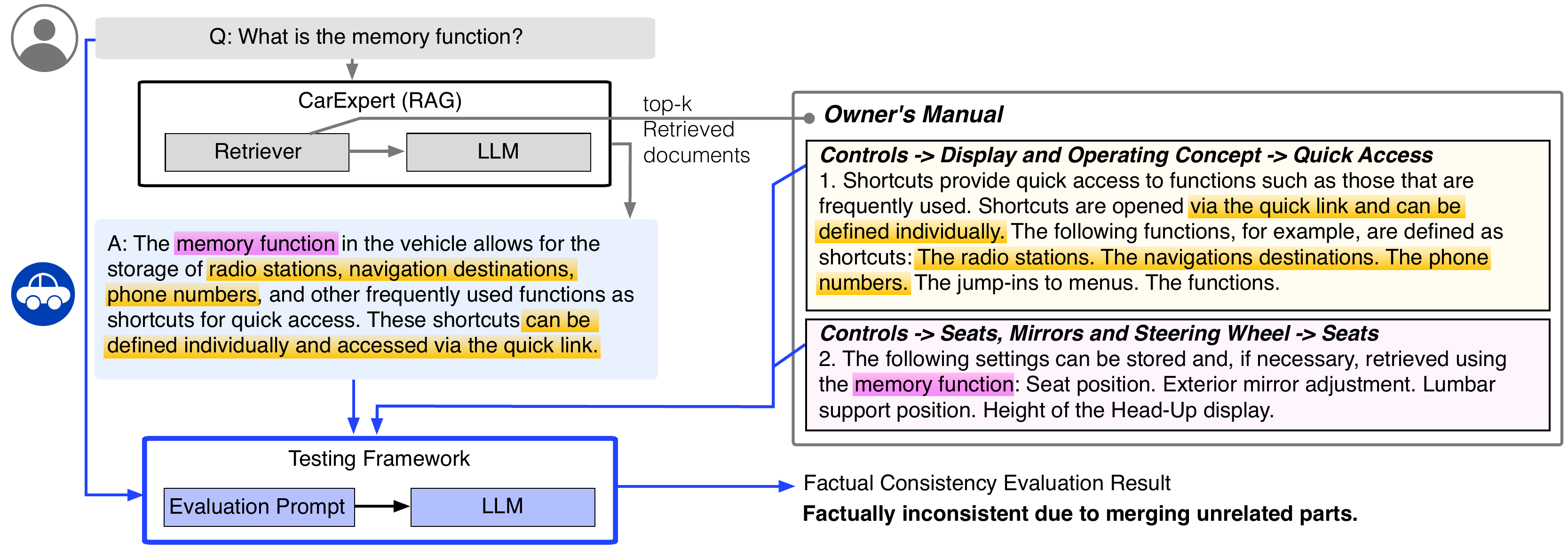}
  \caption{Overall process of our testing framework for the evaluation of factual correctness of in-car conversational systems. 
  } 
  \label{fig:methodology}
\end{figure*}
\section{Testing Conversational Systems}\label{sec:approach}

\subsection{Conversational Question Answering Systems}


At \company, \carqa is an LLM-based question answering system. It can engage with the user in natural, human-like, multi-turn conversations about the car and its functionalities. Its data and ground truth are based on the contents of the owner's manual along with related documents such as press articles. These documents have been parsed and annotated by domain experts, embedded and stored in a vector database.

In this work, we aim to evaluate the answer generation capabilities of \carqa, which we describe next. 
Given a user utterance, \carqa retrieves in-car-domain-specific relevant documents which may include the potential answer. The system uses Retrieval Augmented Generation (RAG) for information retrieval and answer generation. 
Specifically, the modular architecture of \carqa includes four primary sub-components: 1)~orchestration, 2)~semantic search, 3)~answer generation, and 4)~answer moderation. To improve factually correct system utterance generation, \carqa includes control mechanisms on three modules: 1)~it uses an input filter as part of the orchestrator, 2)~it controls the answer generation by prompt design and 3)~it filters it through an output filter. A heuristic is also applied to select the optimal answer, prioritizing both a close match to the user’s question and maintaining as much of the original wording as possible to ensure factual accuracy in information extraction and generation. Further, \carqa integrates text-to-speech and speech-to-text translation, enabling seamless voice-based interaction.
For a comprehensive description, we refer the reader to the relevant literature~\cite{rony-etal-2023-carexpert}.

The use of the retrieved documents for the generation of the output is a critical step which can introduce inconsistencies. For this reason,
a validation approach is required to assess whether generated responses are factually consistent and relevant.
In the rest of this section, we describe our testing methodology to tackle this task.


\subsection{LLM-based Testing Approach}\label{sec:testing-approach}
\autoref{fig:methodology}~(left) shows the high level architecture of our testing method.  The input to our testing method is a user utterance (e.g., a question), whereas the output is a system utterance. The system utterance is represented as textual description of the factual correctness of the original LLM response. The user utterance is passed to \carqa, which retrieves first a list of top-k documents (right boxes) and generates an answer. Then, these artifacts are forwarded with the user utterance to our testing framework to evaluate the factual consistency of \carqa's responses. 

Our testing framework is based on LLMs. In this study we evaluated five distinct LLM types and reasoning approaches, for which we designed customized prompt templates. Examples of prompt templates are given in \autoref{table:prompts_design} (limited to two relevant methods, due to space constraints).


\begin{table}[t]
\scriptsize

\def\arraystretch{1.1}%
\setlength{\tabcolsep}{5.5pt}

\caption{Examples of prompt templates for factual consistency for IO and RT. 
}
\label{table:prompts_design}

\centering

\begin{tabular}{@{}p{0.07\textwidth}p{0.35\textwidth}@{}}

\toprule

\textbf{Method} & \textbf{Prompt Template} \\

\midrule

Input-Output Prompting & 
\ttfamily\makecell[l]{ Setting: a service to help drivers in their \\ car.\\
Consider following sentences:\\ \\ \textcolor{brown}{<retrieved-documents>}.\\ \\
Is the text: \\ \\ \textcolor{blue}{<generated-answer>} \\ \\ factually consistent with the sentences?\\
All information contained in the text has \\ to appear in the sentences.\\ No additional
information must be added\\ or assumed.\\ \\
Return your answer in the following\\ format:\\
\textcolor{red}{<output-format>}.}
\\
\midrule
Round-Table Conference (2nd round only, simplified) & 
\ttfamily\makecell[l]{
Other agents have already returned their\\ evaluation. \\
Check, whether you agree with them or not. \\
\\
Here are their evaluations \\in JSON format:\\
\textcolor{cyan}{<evaluations>}.\\
\\
Return your answer in the following format:\\
\textcolor{red}{<output-format>}.}
\\





\bottomrule
\end{tabular}
\end{table}

In the remaining of this section, we give an overview of the LLM-based methods used in our approach. 

\subsubsection{Input-Output Prompting}

Input-output (IO) prompting, also called \emph{standard prompting}~\cite{wei2023chainofthoughtpromptingelicitsreasoning}, is a straightforward method to query a LLM. A query $x$ is wrapped with predefined task instructions to obtain the system's evaluation~\cite{yao2023treethoughtsdeliberateproblem}.
A more refined version consists in providing examples (few-shot prompting), which has been shown to help the LLM to better understand contextual, domain-specific information, resulting in improved accuracy~\cite{10.1007/978-981-99-8391-9_6}.

\subsubsection{Chain-of-Thought Prompting}

Chain-of-thought (CoT) prompting introduces intermediate thinking steps to better bridge the gap between the input and output. Aside from simple mathematical problems~\cite{yao2023treethoughtsdeliberateproblem}, in practice there is no consolidated definition of what constitutes a step for complex domains such as ours. Nevertheless, CoT has been shown to provide accurate results in several domains~\cite{zhang2022automaticchainthoughtprompting} by providing examples of how a human would sequentially reason about problems. 
In our case, we request in the prompt to iterate over text units of retrieved documents and build up an argumentation chain.

\subsubsection{Self-Consistency with Chain-of-Thought}

Chain-of-though prompting and similar methods create chains autoregressively in one step.
This can lead to error propagation as the number of steps increases due to compound errors~\cite{zhou2024languageagenttreesearch}.
Self-Consistency mitigates this issue, by sampling CoT k-times and then returning the most frequent output.
This has been shown to work if the output space is limited.
If the outputs are too dissimilar, no output would be generated twice (or more)~\cite{yao2023treethoughtsdeliberateproblem}.

\subsubsection{Multi-Persona Self-Collaboration}

Multi-Persona Self-Collaboration (MPSC) dynamically assigns and simulates multiple personas for a task, instructing the LLM to adopt a specific identity. As shown by Xu et al.~\cite{xu2023expertpromptinginstructinglargelanguage}, the restriction to a particular way of thinking may enhance performance.
Personas iteratively generate and critique solutions from other personas, refining them through multiple rounds to achieve consensus.
This method, called Solo Performance Prompting (SPP), operates in a pure zero-shot manner \cite{wang2024unleashingemergentcognitivesynergy}.
In our framework we defined first the personas \textit{Fact Checker, Research Analyst, Editor, Journalist, Librarian} and questioned the LLM Claude 3 Haiku~\cite{TheC3} do describe the personas. We assume, that these personas possess qualifications which are necessary for a critical review of factual consistency.




\subsubsection{Round Table Conference}

While previous methods rely on a single large language model to complete the task, using multiple LLMs---often referred to as agents---offers the potential for even better performance. The approach involves allowing each agent to independently generate its own reasoning and then engage in a collaborative discussion until reaching a consensus.

We use a method similar to ReConcile~\cite{chen2024reconcileroundtableconferenceimproves}, a Round Table (RT) Conference approach, which adds additional weighting of the responses, based on the returned confidence estimations of the agents, as well as more elaborate examples.
If all agents agree on a common evaluation, or if the maximum number of rounds has been reached, the Round Table Conference ends and returns the common answer, otherwise the next round is initiated.
In our preliminary experiments, we determined that a limit of five agents was sufficient to produce reliable results, which is why we selected it as our threshold.
\section{Case Study at \company}
\label{sec:empirical-study}

\subsection{Research Questions}
\label{sec:rqs}

We consider the following research questions.

\noindent
\textbf{RQ\textsubscript{1} (effectiveness):} \textit{How accurate is our framework in assessing factual relevance and consistency?}



\noindent
\textbf{RQ\textsubscript{2} (efficiency):} \textit{How efficient is our framework?}

The first research question investigates the effectiveness of our framework. We evaluate different configurations and influencing factors such as the LLM instance, method or temperature used. The second research question targets the evaluation of the efficiency of different methods to investigate which methods are suitable for real-time processing.

\subsection{Metrics}
\label{sec:metrics}



To evaluate the effectiveness of our framework (\textbf{RQ\textsubscript{1}}), we compare the factual relevance and consistency evaluation results output by our framework with an expert-based evaluation. Specifically, we assess the \textit{accuracy} between the evaluation of the human and our proposed framework (as outlined in (1)), similarly to previous studies on the performance of LLM-based systems~\cite{honovich2022truereevaluatingfactualconsistency}. 

\begin{align}
    \textit{ACC} = \frac{\text{\# agreements expert and our framework}}{\text{\# samples}}
\end{align}

We apply the metric respectively for both, the factual relevance and consistency dimension.
To date, we consider our choice as the best available strategy for evaluating our approach for the following reasons: 1) conventional automated evaluation metrics such as ROUGE, BLEU or other n-gram based metrics fail to accurately analyze semantics, which is the focus of our study. 2) conventional metrics often exhibit a low correlation with human judgments~\cite{liu2023gevalnlgevaluationusing}. 
A description of how the factual relevance and consistency labels are created by an expert is outlined in the following \autoref{sec:dataset}.

For \textbf{RQ\textsubscript{2}}, we recorded the execution time for each request and identified the consumed token number including the number of tokens in the prompt for the evaluation task and the response generated by our system.




\subsection{Dataset Creation}
\label{sec:dataset}


We use the owner's manual as our ground truth. First, we parse the owner's manual of a specific \vehiclebrand into a JSON file and divide its content into paragraphs. We call these \textit{documents}.
Overall, we retrieve a total of 4914 documents.
We then let \company experts go through the documents and manually create questions, i.e., user utterances. For each question, there is exactly one document sufficient to address the question. Then, we query \carqa with the  questions from the experts and receive responses, including a list of retrieved documents and the generated answer to our question. After this, we pass the question with the outputs from \carqa to two different experts, both employed in the Car Manual Quality Assurance departement with more then 5 years experience to decide whether the generated answer is 1)~consistent with the retrieved document(s), and 2)~whether it is relevant. In particular, to achieve a reliable dataset, we randomly repeat questions passed to the first reviewer. If the expert provides inconclusive labeling results when questioned multiple times, we pass the evaluation once to the second expert to decide on the label. This process generates us consistency and relevance labels for each question and answer pair, which we consider as our ground truth.

\subsection{Configurations}
\label{sec:comfigurations}

We evaluate our framework with five different proprietary and open-weight models, also considering their size in terms of number of parameters. 
Particularly, for our experiments, we chose widely popular models from OpenAI, such as GPT-3.5-turbo, GPT-4 and their to date latest model GPT-4o. As a representative of open models, we adopted two versions of Meta Llama 3, with 8 billion and 70 billion parameters.

For all LLM-based methods and types as described in \autoref{sec:testing-approach}, we executed the experiments varying  four different temperature settings in the range  $[0.0, 0.2, 0.4, 0.6]$. The pre-defined prompts were kept the same for all experiments conducted. Overall, we performed $10,300$ evaluations ($103$ samples $\times$ $5$ LLM-based methods $\times$ $5$ LLMs $\times$ $4$ temperatures).
\subsection{Implementation}
Our benchmarking framework is implemented in Python and is using the LangChain\footnote{https://www.langchain.com/} library to send evaluation requests, containing the initial users utterance and the output of \carqa, to the considered LLMs. \carqa is deployed locally, while the to be tested LLMs for our evaluation are deployed in the Cloud. Specifically, the LLAMA models are deployed in Azure Virtual Machines, while GPT-based models are hosted natively by OpenAI.
For the expert evaluation and ground truth data creation we create an online application, where the expert is presented the original user utterance,  the answer by \carqa and retrieved document. The accuracy evaluation comparing the experts results with the ones from our framework is performed locally.

\begin{table}[t]

\centering

\caption{RQ\textsubscript{1}: Accuracy results of various LLM and method combinations, evaluated by experts for relevance and consistency across different models and temperature settings (percentage-based). The Pareto-optimal solutions, which are not simultaneously dominated in both relevance and consistency, are highlighted in bold.}
\label{table:rq1234}

\setlength{\tabcolsep}{5pt}
\renewcommand{\arraystretch}{1.1}

\begin{tabular}{@{}lllllllll@{}}

\toprule

& \multicolumn{4}{c}{\sc Relevance} & \multicolumn{4}{c}{\sc Consistency} \\ 



\cmidrule(r){2-5}
\cmidrule(r){6-9}

& 0.0 & 0.2 & 0.4 & 0.6 & 0.0 & 0.2 & 0.4 & 0.6 \\ 

\midrule

\textbf{GPT-3.5-turbo} \\ [0.1em]

\quad RT & \textbf{93.2} & \textbf{93.2} & 92.2 & 92.2 & \textbf{21.3} & \textbf{20.3} & 20.3 & 21.3 \\
\quad CoT & 90.2 & 89.3 & 88.3 & 88.3 & 22.3 & 25.2 & 31.0 & 29.1 \\
\quad CoT-SC & 90.2 & 89.3 & 90.2 & 89.3 & 22.3 & 22.3 & 24.2 & 23.3 \\
\quad IO & 90.2 & 89.3 & 92.2 & 89.3 & 82.5 & 81.5 & 76.6 & 69.9 \\
\quad MPSC & 90.2 & 91.2 & 93.2 & 90.2 & 72.8 & 74.7 & 72.8 & 69.9 \\ [0.5em]


\textbf{GPT-4} \\ [0.1em]

\quad RT            & \textbf{92.2} & 92.2 & \textbf{92.2} & \textbf{92.2} & \textbf{90.2} & 89.3 & \textbf{90.2} & \textbf{91.2} \\
\quad CoT                 & 92.2 & 91.2 & 90.2 & 92.2 & 89.3 & 88.3 & 88.3 & 90.2 \\
\quad CoT-SC        & 92.2 & \textbf{92.2} & 91.2 & 91.2 & 89.3 & \textbf{89.3} & 89.3 & 89.3 \\
\quad IO                     & \textbf{90.2} & \textbf{91.2} & 89.3 & 90.2 & \textbf{92.2} & \textbf{91.2} & 90.2 & 91.2 \\
\quad MPSC & 82.5 & 77.6 & 77.6 & 78.6 & 82.5 & 87.3 & 85.4 & 84.4 \\ [0.5em]


\textbf{GPT-4o} \\ [0.1em]

\quad RT            & 91.2 & 89.3 & 90.2 & 90.2 & 85.4 & 84.4 & 87.3 & 85.4 \\
\quad CoT                 & 78.6 & 80.5 & 79.6 & 80.5 & 65.0 & 65.0 & 64.0 & 64.0 \\
\quad CoT-SC        & 78.6 & 79.6 & 77.6 & 79.6 & 65.0 & 66.9 & 65.0 & 57.2 \\
\quad IO                     & 78.6 & 78.6 & 81.5 & 78.6 & 79.6 & 79.6 & 77.6 & 78.6 \\
\quad MPSC & 70.8 & 71.8 & 71.8 & 71.8 & 78.6 & 80.5 & 78.6 & 80.5 \\ [0.5em]


\textbf{LLAMA 3 8B} \\ [0.1em]

\quad RT            & 14.5 & 15.5 & 14.5 & 15.5 & 82.5 & 82.5 & 82.5 & 84.4 \\
\quad CoT                 & 85.4 & 85.4 & 89.3 & 86.4  & 52.4 & 79.6 & 76.6 & 73.7 \\
\quad CoT-SC        & 85.4 & 85.4 & 86.4  & 83.4 & 79.6 & 79.6 & 72.8  & 76.6 \\
\quad IO                     & 85.4 & 86.4  & 85.4 & 83.4 & 49.5 & 71.8 & 75.7 & 64.0 \\
\quad MPSC & 66.9 & 66.0 & 72.8  & 61.1 & 55.3 & 67.9 & 64.0 & 66.0 \\ [0.5em]


\textbf{LLAMA 3 70B} \\ [0.1em]

\quad RT            & 91.2 & 91.2 & 91.2 & 91.2 & 81.5 & 22.3 & 20.3 & 22.3 \\
\quad CoT                 & 91.2 & 91.2 & 91.2 & 92.2 & 84.4 & 84.4 & 84.4 & 85.4 \\
\quad CoT-SC        & 90.2 & 91.2 & 91.2 & 92.2 & 84.4 & 85.4 & 85.4 & 84.4 \\
\quad IO                     & 91.2 & 91.2 & 91.2 & 91.2 & 83.4 & 83.4 & 83.4 & 84.4 \\
\quad MPSC & 82.5 & 82.5 & 82.5 & 84.4 & 82.5 & 85.4 & 83.4 & 85.4 \\ 

\bottomrule

\end{tabular}
\end{table}

\subsection{Results}

\autoref{table:rq1234} reports the results for our research questions RQ1. The table presents the accuracy results, for different temperatures, LLMs, and evaluation methods, for both relevance and consistency. As it is challenging to compare methods whose performance is expressed by multiple conflicting metrics/objectives, i.e., relevance and consistency in our case, we have sorted the results based on Pareto-dominance and highlighted the results in boldface. Nevertheless, the use of our benchmarking framework can also target specifically relevance or consistency, which is why we report in addition our observations and analysis for each metric independently. 


\head{RQ\textsubscript{1}}
Regarding relevance, the highest score is achieved with GPT-3.5-Turbo with 93.2\% at temperature 0, while the corresponding consistency value was around 20\%. However, GPT-4 achieves with RT a slightly lower relevance accuracy of 92.2\% but a significantly higher consistency score of 90.2\% at the same temperature.
We encountered the exact opposite result with GPT-4 using IO. 
Solely results from GPT-3.5 Turbo and GPT-4 based methods pose the set of non-dominated results, where GPT-4 scores do not fall below 89.3\%, which is the worst score for consistency at temperature 0.2. 

\textbf{Impact of different LLMs/methods.} For GPT-4 we did not observe a significant impact of the evaluation method used, only MPSC performed worse for almost all temperatures then the best results of GPT-4 for both relevance and consistency. For GPT-3.5 turbo we observed less variance regarding relevance. However, results diverge more for consistency, which is also the case for GPT-4o and LLAMA 3 8B. For instance, RT exhibits a low consistency score of 21.3\% at temperature 0 and 22.3\% for GPT-3.5-turbo and LLAMA 3 70B, while methods such as IO achieve scores over 80\%.

Regarding consistency, the methods' impact depends on the LLM used. For instance, at temperature 0, the methods CoT, Cot-SC exhibit for GPT-3.5-turbo scores around 20\%, while for other LLMs the results are significantly better. Another example is the approach MPSC, which achieved the best consistency scores for GPT-4, GPT-4o and LLAMA 3 70B, while showing the lowest score for Llama 3 8B with 55.3\% at temperature 0.





\textbf{Impact of different temperatures.} Regarding the impact of the temperature on the results, for relevance we did not observe a significant variation when increasing the temperature. The highest decrease was for LLAMA 3 8B with MPSC by 5.8\%, from temperature 0 to temperature 6.

For consistency, we observed the highest decrease from 81.5\% to a round 22\% with LLAMA 3 70 B and RT. The highest increase was observed for LLAMA 3 8B for CoT from 52.4\% to over 73\%. For the other LLMs and methods the scores tend slightly to decrease or increase without a concrete trend.


\begin{table}[t]
    \centering 
  \caption{Scores for Round Table Conferences with agents consisting of different LLMs with a temperature of 0.0.}
  
  \setlength{\tabcolsep}{10pt}
\renewcommand{\arraystretch}{1.1}

  \label{table:multiple_LLM}
\begin{tabular}{@{}lll@{}}
\toprule
\textbf{Round Table Agents}                        & \textbf{Consistency} & \textbf{Relevance} \\ \midrule
5x GPT-4                               & \textbf{90.2}        & \textbf{92.2}      \\
1x GPT-4, 1x LLAMA 3 70B            & 89.3        & 90.2      \\
1x GPT-4, 1x GPT-4o, 1x LLAMA 3 70B & 83.4        & 90.2      \\
2x GPT-4, 1x GPT-4o, 2x LLAMA 3 70B & 85.4        & 89.3      \\ \bottomrule
\end{tabular}
\end{table}

\textbf{Multi-Agent Round Table.}
The former Round Table conference approach used multiple instances of the same LLM. In additional, we evaluated Round Table based reasoning using different agents/LLMs.
The idea of using various agents is to encourage divergent thinking, which is expected to improve the results.

We performed additional experiments with the RT approach by combining several of the top-performing LLMs (GPT-4, GPT-4o, and LLAMA 3 70B) in various ensemble configurations at temperature 0. The results, as shown in \autoref{table:multiple_LLM}, however indicate that an ensemble of 5 GPT-4 models achieved the highest relevance and consistency scores, which is similar to the performance of a single GPT-4 model employing RT. Combining agents of different types yielded lower relevance and consistency scores, achieving a higher decrease in consistency than for relevance.




\begin{tcolorbox}
\textbf{RQ\textsubscript{1} (effectiveness):} 
GPT-4 balanced relevance and consistency best, scoring 92.2\% vs. 90.2\% with RT and 90.2\% vs. 92.2\% with IO at temperature 0, with minimal variation across methods. GPT-3.5-Turbo with RT achieved the highest relevance (93.2\%), while LLAMA 3B had the lowest (14.5\%, with 82\% consistency). Generally, relevance scores surpassed consistency, which varied by LLM. GPT-3.5-Turbo's CoT and CoT-SC scored around 20\%, whereas other LLMs performed better. Temperature changes had little impact, and using multiple agents for GPT-4 RT reduced relevance more than consistency.
\end{tcolorbox}

\begin{table*}[t]
    \centering
     \caption{Comparison of Owner's Manual Text and Generated Answers for examples of three categories of error types.}
    \label{tab:qualitative}
    \begin{tabular}{c  p{0.9\textwidth}}
    \toprule
    Type 1 & \textbf{Manual:} {\ttfamily To manually turn on \colorbox{green!30}{standby} state, press and hold the thumbwheel on the center console.} \\ 
     & \textbf{Generated:} {\ttfamily To manually turn on \colorbox{blue!30}{idle} state, press and hold the thumbwheel on the center console.} \\ 
    \midrule
    Type 2 & \textbf{Manual:} {\ttfamily Press and hold the unlock button on the vehicle key after unlocking. The windows open for as long as the button on the vehicle key is pressed.} \\ 
    &\textbf{Generated:} {\ttfamily \colorbox{blue!30}{The welcome window} is a feature that allows you to open the windows of the vehicle by pressing and holding the unlock button on the vehicle key after unlocking. [...]} \\ 
    \midrule
    Type 3 & \textbf{Manual:} {\ttfamily The Lane Change Warning system with active return detects vehicles in your blind spot or vehicles approaching from behind in the adjacent lane.} \\ 
    & \textbf{Generated:} {\ttfamily The Lane Change Warning light illuminates when there is a risk of collision with a vehicle crossing from the left [...] and \colorbox{blue!30}{immediate} \colorbox{blue!30}{braking} or an evasive maneuver is required.} \\ 
    \bottomrule
    \end{tabular}
\end{table*}

\begin{table}
\centering
\caption{Overview of execution times, token number for one evaluation, and average time per token.}
\label{tab:efficiency}
\begin{tabular}{lrrc}
\toprule
\textbf{Method} &  \textbf{Tokens} &  \textbf{Execution Time (s)} & \textbf{Average Time (s/token)} \\
\midrule
    RT &    3924 &                23.7 &                      0.0060 \\
  MPSC &    2376 &                17.8 &                      0.0074 \\
   CoT &    1427 &                 5.5 &                      \textbf{0.0038} \\
CoT-SC &    4281 &                16.5 &                      0.0039 \\
    IO &     \textbf{689} &                 \textbf{4.5} &                      0.0065\\
\bottomrule
\end{tabular}
\end{table}

\head{RQ\textsubscript{2}}
\autoref{tab:efficiency} reports the average token usage per evaluation and average time per evaluation for methods on GPT-4.
Since LLAMA models were deployed on Azure virtual machines, a direct cost and time comparison with GPT models is not feasible. OpenAI’s models, being proprietary and available only as serverless functions, use a distinct pricing and compute structure. We assume that the relative differences between methods are consistent for LLAMA models as well. The best efficiency regarding tokens per time achieved COT, followed by COT-SC and RT. Regarding the evaluation time for one request, IO was the fastest method.

\begin{tcolorbox}
\textbf{RQ\textsubscript{2} (efficiency):} 
The most efficient method in terms of time required per token was CoT (6.5ms/token), followed by CoT-SC and RT. However, when considering the evaluation time for a single request, IO demonstrated the best performance with 4.5s per request compared to 23.7s for CoT.


\end{tcolorbox}



\subsection{Qualitative Analysis}


We analyzed our results to identify specific failure patterns and gain insight as of why our methods did not achieve 100\% accuracy for factual relevance and consistency.
However, a manual error analysis of the over 23,000 individually generated evaluations would be infeasible, thus we only analyzed the results of the best run of each of the methods.
Since relevance tends to be assessed on an intuitive basis, we focus our analysis on factual consistency, as this is based on facts and can therefore be analyzed more objectively.
An error is identified when the expert opinion was positive but the implemented methods returned negative results, or vice versa.
We ignore errors regarding failed requests, formatting difficulties as they only refer to errors in the results themselves.
We identified three categories of errors~\cite{2020-Humbatova-ICSE}, described next.
None of the LLM techniques used in our framework succeeded in adequately answering the questions for the examples in Table~\ref{tab:qualitative}.

\subsection*{Error Type 1: Confusion of Internal Terminology}

One example for error type 1 is the response to a user utterance about the ``standby state'' as given in Table~\ref{tab:qualitative}. \carqa used information about the ``idle state”, although this is a different state. The evaluation stated the statement as factual consistent, although the statement is incorrect. 








Without a domain-specific understanding of these terms, completing this request is challenging for the system. During LLM model training, terms like ``standby'' and ``idle'' may have be treated as synonyms. However, within a brand-specific context, these terms represent distinct functions. LLMs may also struggle with the company's evolving terminology, as new functions and names are regularly introduced. 


\subsection*{Error Type 2: Hallucinations}

Error of type 2 involves hallucinations. Hallucinations occur when \carqa attempts to respond despite retrieving documents from the user manual that either did not address the question or provided inadequate answers.
For instance, a question regarding the ``welcome window'' as shown in Table~\ref{tab:qualitative} led to an imagined connection with a retrieved document from the owner's manual, incorrectly associating it with the requested function. 







The ``welcome window'' is a small greeting message on the infotainment, but the \carqa system used other information here and interpreted that the window in the given extract from the owner's manual is the `welcome window', although not referenced.

\subsection*{Error Type 3: Common Sense Errors}

The final type of error involves a lack of common sense. While it could be considered as a subset of hallucinations, we distinguish it because additional and specialized knowledge the mistake is unlikely to be made by a human. As an example, consider the term ``Lane Change Warning light''. As the name implies, this light alerts the driver to a lane change if there is already a vehicle in an adjacent lane.








While the system's purpose is obviously clear, the \carqa's response---to recommend immediate braking---is wrong and safety-critical.






\section{Related work}\label{sec:related-work}

Several papers have already outlined LLM-based evaluation methods~\cite{chiang2023largelanguagemodelsalternative, lin2023llmevalunifiedmultidimensionalautomatic, honovich-etal-2021-q2} and its advantages~\cite{hosking2024humanfeedbackgoldstandard}, but these work focus on text summarization. In contrast, our method focuses on evaluating in-car conversational systems in the automotive domain for factual correctness.

Recently, researchers have proposed using LLMs to evaluate other LLMs~\cite{zheng2023judgingllmasajudgemtbenchchatbot} (i.e., ``LLM-as-a-Judge'').
The setup includes several LLMs, which generate answers to predefined, open-ended questions.
After this. experts are used to evaluate and label the generated answers.
The LLMs judges are then given the same question-answer pairs, which they will also have to evaluate on their own.
In this experiment the LLM judgements reached an agreement of over 80\% with humans, proving a promising way of evaluating LLM generated answers.
Explicitly analyzing the factual correctness and consistency of the generated output has been neglected, going even as far as mentioning safety has been disregarded in the analysis, which is a requirement for in-car usage~\cite{zheng2023judgingllmasajudgemtbenchchatbot}.

In the automotive domain, Friedl et al.~\cite{friedl2023incarethinkingincarconversational} describes the evaluation of in-car conversational systems using LLMs. Multiple personas were created to let the LLM judge the quality of the generated output. However, the evaluation only included metrics like ``follow-up'', ``implicit understanding'' and ``harmful user input'', neglecting yet again factual correctness, factual consistency and relevance of the answer~\cite{friedl2023incarethinkingincarconversational}. 
Our work, on the other hand, focuses on the correctness of the main functionality of conversational systems, thus it complements existing research.

\section{Conclusions}\label{sec:conclusions}

In this work, we have proposed a validation approach using LLMs to test an in-car conversational system, focusing on factual correctness, spanning the dimensions of factual relevance and factual consistency. We have created a human-based curated dataset specific for the in-car domain to evaluate our approach.
Our evaluation of five advanced LLM types and methods shows that multiple models achieve over 90\% accuracy in relevance and consistency across various hyperparameter settings.
The best results regarding the tradeoff between relevance, consistency, and efficiency where achieved with GPT-4 employing the Input-Output prompting. 
Overall, our study shows the potential of LLM-based methods for the automated evaluation of factual correctness in conversational question answering systems and intends to encourage further studies.

In our future work we want to investigate approaches for the synthetic data generation creating pairs of questions and source documents, thereby avoiding expensive expert-crafted questions.
Further, we will investigate the performance of our approach when employing other languages or vehicle types.








\balance
\bibliographystyle{IEEEtran}
\bibliography{paper.bib}

\end{document}